\pdfoutput=1

\documentclass[11pt]{article}

\usepackage[]{ACL2023}

\usepackage{times}
\usepackage{latexsym}

\usepackage[T1]{fontenc}

\usepackage[utf8]{inputenc}

\usepackage{microtype}

\usepackage{booktabs}
\usepackage{bm}
\usepackage{adjustbox}
\usepackage{multirow}
\usepackage{lipsum}
\usepackage{amsfonts}
\usepackage[inline]{enumitem}
\usepackage{amsmath}
\usepackage{url}
\usepackage{xspace}

\newcommand{\hit}{iHT\xspace}
\newcommand{\hitctx}{iHT$_{\mathcal{G}}$\xspace}
\newcommand{\hitent}{iHT$_{\mathcal{E}}$\xspace}
\title{Pre-training Transformers for Knowledge Graph Completion}

\author{
 Sanxing Chen$^\spadesuit$\thanks{\hspace{.06in}Part of this work was done during an internship at Microsoft Research.}\quad Hao Cheng$^\ddagger$\quad Xiaodong Liu$^\ddagger$\quad Jian Jiao$^\dagger$\\\textbf{Yangfeng Ji$^\clubsuit$\quad Jianfeng Gao$^\ddagger$} \\
 $^\spadesuit$Duke University \quad
 $^\ddagger$ Microsoft Research \quad $^\dagger$Microsoft Bing Ads \quad $^\clubsuit$University of Virginia \\
 \texttt{sanxing.chen@duke.edu} \\
 \texttt{\{chehao,xiaodl,jiajia,jfgao\}@microsoft.com} \\
 \texttt{yangfeng@virginia.edu}
}

\begin{document}
\maketitle
\begin{abstract}
Learning transferable representation of knowledge graphs (KGs) is challenging due to the heterogeneous, multi-relational nature of graph structures.
Inspired by Transformer-based pretrained language models' success on learning transferable representation for texts, we introduce a novel inductive KG representation model (\hit) for KG completion by large-scale pre-training.
\hit consists of a entity encoder (e.g., BERT) and a neighbor-aware relational scoring function both parameterized by Transformers.
We first pre-train \hit on a large KG dataset, Wikidata5M.
Our approach achieves new state-of-the-art results on matched evaluations, with a relative improvement of more than $25\%$ in mean reciprocal rank over previous SOTA models.
When further fine-tuned on smaller KGs with either entity and relational shifts, pre-trained \hit representations are shown to be transferable, significantly improving the performance on FB15K-237 and WN18RR.

\end{abstract}

\section{Introduction}
As a fundamental component of human intelligence, relational knowledge plays a crucial role in imitating human cognitive abilities with machine learning~\cite{halford2010relational}.
Knowledge graphs (KGs) are the most widely used representation of relational knowledge, with well-known examples such as Freebase~\cite{bollacker2008freebase}, YAGO~\cite{suchanek2007yago}, and Wikidata~\cite{vrandevcic2014wikidata}.
KG is also a key ingredient for many natural language understanding tasks, such as language modeling~\cite{peters-etal-2019-knowledge}, question answering~\cite{sun-etal-2018-open}, and commonsense reasoning~\cite{bosselut-etal-2019-comet}.
Despite significant human efforts in constructing KGs, they are still far from being complete, which motivates the development of automatic KG completion models.

A long line of past work has tried to explore generalizable representation for KG completion.
Conventional embedding approaches focus on transductive generalizability, i.e., learning entity representation from the structure of KGs to infer new facts among existing entities~\cite{wang2017knowledge}.
On the other hand, inductive approaches have also been studied to accommodate new entities that are not present in the training KGs, e.g., by building compositional representation of entities based on their textual descriptions~\cite{wang-etal-2014-knowledge,xie2016representation}.
However, as the research in both direction only leverages in-domain data, the knowledge obtained is limited within a fixed body of KG data that are available in the domain of interest.

Inspired by the success of transfer learning in natural language processing (NLP), recent work exploits knowledge stored in large amount of unstructured texts, leveraging large-scale pre-trained language models (LMs) to improve the performance of KG completion models~\cite{yao2019kgbert,clouatre-etal-2021-mlmlm}.
Co-training LMs and KG completion models has been shown to be effective in improving the performance of downstream knowledge-intensive NLP tasks, but not so much for the KG completion task itself~\cite{wang2021KEPLER,yasunaga2022deep}.
Despite the progress on transferring knowledge between structured KGs and unstructured texts, the generalization from one KG to another is still an open problem that is rarely studied~\cite{kocijan-lukasiewicz-2021-knowledge}.

\begin{figure*}[tb]
  \centering
  \includegraphics[width=\linewidth]{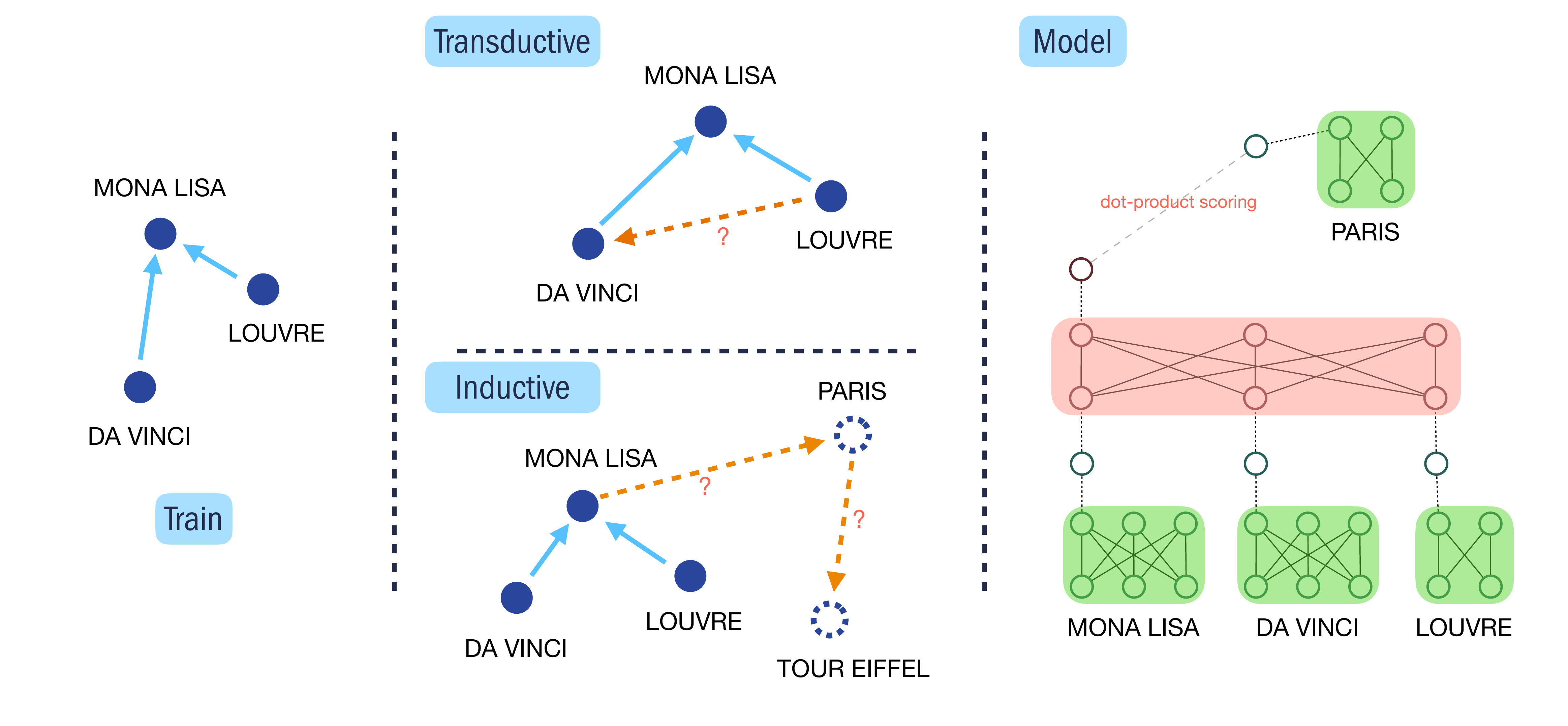}
  \caption{Inductive KG completion requires relation prediction involving unseen entities (e.g., \emph{Paris} and \emph{Tour Eiffel}).
  \hit employs an entity Transformer (bottom green blocks) to encode entity surface forms to entity representations and a context Transformer (top red blocks) to model (optional) multi-relational graph contexts and relational queries.
  Large-scale pre-trained \hit can be continually trained on other KGs.
  }
  \label{fig:model}
\end{figure*}

In this paper, we propose to pre-train a Transformer-based inductive KG representation model for KG completion.
Our goal is to learn transferable knowledge representations with richer semantics from both unstructured texts (by initialization from pre-trained LMs) and structured KGs (by pre-training on a large-scale KG, Wikidata5M).
We first introduce \hit as our pre-training backbone.
\hit employs the typical Transformer encoder for processing entity surface forms and is augmented with a Transformer-based relational scoring module~\cite{chen-etal-2021-hitter}, which is proven to be more expressive and versatile than conventional approaches, to capture various KG contexts.
The hope is, via a unified Transformer architecture that underlies recent successes in transfer learning, the pre-trained model can be continually adapted to different KGs and downstream tasks.

We pre-train \hit on the Wikidata5M dataset, which is an encyclopedic KG containing millions of entities and triplets for KG completion and show that \hit outperforms previous state-of-the-art approaches by more than 25\% relatively in terms of MRR.
Upon further analysis, we carefully examine several critical design choices and discuss important factors such as model parameterization and negative sample size in building large-scale knowledge representation models for pre-training.
We then test \hit on two standard KG completion benchmarks (FB15K-237 and WN18RR) by continual training 
 the pre-trained model and show that large-scale pre-training significantly improves the performance of \hit on KGs of different domains.

\section{\hit}
Similar to conventional knowledge graph embedding (KGE) approaches, we train \hit on the KG completion task.
Given a KG representing as a set of fact triplets ($\mathcal{G}=\{(e_s, r_p, e_o), \dots\}$), where each triplet consists of two entities (subject entity $e_s$ and object entity $e_o$) and a predicate $r_p$, the task basically holds out one of the entities and requires the model to recover it based on the rest information.
For example, without loss of generality, when $e_o$ is missing the model will be using $(e_s, r_p, )$ as input to recover it in the output.

Our proposed model has two major components, i.e., an entity encoder to map the surface form of an entity to a continuous embedding space; and a context encoder to contextualize the entity embedding with a relational query and the entity's graph neighborhood.
We implement both encoders using the Transformer architecture with the hope to take advantage of the architecture's scalability and transferability as shown in pretrained language models.
We detail their design in this section.

\subsection{Entity Transformer}
Conventional knowledge embedding approaches treat an entity as the atomic unit. As a result, the parameter size increases linearly with the number of entities. Scaling up to large-scale KGs with millions of entities in this way is challenging and computationally inefficient.
Moreover, due to this transductive setting assuming all entities are seen during training, such approaches can not adapt to new entities.

However, entities in KGs of common domains such as encyclopedia are often associated with surface forms (e.g., entity description, name and alias).
These textual representations leverage compositional semantics to depict any entity with a small vocabulary (typically tens of thousands) in natural language, connecting the textual semantics of an entity to its symbolic representation.
If we can learn an entity encoder to uniquely identify an entity based on its surface forms, we can not only solve the scalability problem but also can potentially enable compositional knowledge sharing across entities.
Our entity Transformer is designed to achieve this goal with a BERT-like encoder.

Specifically, we follow the same input format as specified in \citet{devlin-etal-2019-bert}, we tokenize the entity's surface form into subwords $e=(e^1, \dots, e^L)$ and pad the subword sequence with special tokens.\footnote{Position embeddings and type embeddings (one type) are also added as in BERT.}
\begin{equation}
\mathtt { Seq }_{\text {BERT }}(e) = \texttt{[CLS]}\xspace e \texttt{[SEP]}
\end{equation}

\begin{table*}[h]
\centering
\begin{adjustbox}{max width=0.9\linewidth}
\begin{tabular}{llrrrrrr}
\toprule
                       &          & \multicolumn{3}{c}{Transductive} & \multicolumn{3}{c}{Inductive}   \\\cmidrule(r{4pt}){3-5}\cmidrule(r{4pt}){6-8}
                       &          & Wikidata5M  & FB15K-237 & WN18RR & Wikidata5M & FB15K-237 & WN18RR \\\midrule
\multirow{2}{*}{Train} & relation &    822         &     237      &    11    &      822      &     237      &    11    \\
                       & entity   &    4,594,485         &   14,505        &    40,559    &    4,579,609        &     11,633      &    32,755    \\
                       & triplet  &    20,614,279         &   272,115        &   86,835     &    20,496,514        &    215,082       &    69,585    \\\midrule
\multirow{3}{*}{Dev}   & relation &    822         &    223       &    11    &       199     &      225     &    10    \\
                       & entity   &    4,594,485         &   9,809        &   5,173     &   7,374         &        8,965   &     9,736   \\
                       & triplet  &    5,163         &   17,535        &   3,034     &     6,699       &     42,164      &    11,381    \\\midrule
\multirow{3}{*}{Test}  & relation &    822         &     224      &    11    &      201      &     232      &    10    \\
                       & entity   &    4,594,485         &    10,348       &   5,323     &   7,475         &      10,645     &    10,223    \\
                       & triplet  &    5,133         &    20,466       &   3,134     &     6,894       &      52,870     &    12,037   \\\bottomrule
\end{tabular}
\end{adjustbox}
\caption{Statistics of three knowledge graph completion datasets. Numbers of entities in inductive training sets are generally smaller than those under the transductive setting, meaning more unseen entities in their dev and test sets.}
\label{tab:datasets}
\end{table*}

We take the final hidden state corresponding to the \texttt{[CLS]} token from the entity Transformer as the entity embedding.
 
\begin{equation}
\mathtt { Embed }_{\text {ENT }}(e)=\mathtt{ET}_{\text {[CLS] }}\left(\mathtt { Seq }_{\text {BERT }}(e)\right)
\end{equation}
As our entity Transformer shares the same architecture with BERT-like language models, it is natural to use pretrained language model weights to initialize the entity Transformer.

\subsection{Context Transformer}

Recent Transformer-based approaches for KG completion often employ relational scoring functions from conventional KG embedding approaches, such as TransE~\cite{bordes2013transe} and DistMult~\cite{yang2014distmult}.
However, these simple and often non-trainable scoring functions can only incorporate the relational query, ignoring rich graph contexts in KGs and leaving all the heavy lifting in learning and storing relational knowledge to the entity encoder.
We instead draw on recent successes in using Transformer as both contextualizer and relational scorer~\cite{chen-etal-2021-hitter} to build our context Transformer.

There are two major types of contexts that we want to incorporate, i.e., the relational query $r_p$ and a set of directly linked entities $N_G(e_{s})=\{(r_1, e_{n_1}), \dots, \}$ in the neighborhood of the source entity, where $r_i$ is the relation type between $e_{s}$ and $e_{n_i}$.
We use the summation of the embeddings of neighborhood entity $\mathtt { Embed }_{\text {ENT }}(e_{n_i})$ and the embeddings of the corresponding relation type as a relation-dependent entity representation for each neighborhood entity.\footnote{As noted in \citet{wang2021KEPLER}, the relation descriptions in Wikidata5M are unspecific, which leads to worse performance, thus we use lookup embeddings for relation types. Future work can replace them with compositional representations.}
These relation-dependent entity representations are concatenated with the relational query $r_p$, the source entity, and a special \texttt{[GCLS]} token as the inputs to the context Transformer.
We also add type embeddings to distinguish different types of inputs, which eliminates the need for additional position embeddings.
The output corresponding to the \texttt{[GCLS]} token can be seen as a context-dependent entity embedding for the source entity $\mathtt { Embed }_{\text {CTX }}(e_{s})$.

\subsection{Link Prediction}
We train the \hit model with a link prediction objective.
Given the positive $(e_s, r_p, e_o)$ triplet, we sample a set of target entities uniformly to construct negative triplets.
Empirically we find that model performance improves with the increasing number of negative samples.
In order to achieve the largest possible negative sample size under memory constraints, we use in-batch negative sampling where negative targets are shared across all examples in the same batch and encoded only once.
We reuse the entity Transformer to get the embeddings of every negative target entity and the only positive target entity $e_o$.
We then compute the dot-product similarity between each of them and $\mathtt { Embed }_{\text {CTX }}(e_{s})$.
The loss is defined as the cross entropy between the softmax normalized distribution and the ground truth one-hot distribution.
During the inference stage, such link prediction task is performed over all possible triplets.

\section{Experiments}

In this section, we examine the ability of \hit to handle a large-scale KG and generalize \emph{in-domain} to unseen facts of known entities (transductive setting) and unseen facts with unseen entities (inductive setting).

\subsection{Datasets}

Table~\ref{tab:datasets} shows the statistics of the three datasets used in our experiments.
The experiments of this section are conducted on the Wikidata5M dataset~\cite{wang2021KEPLER}, which is the largest of the three.
The dataset, based on Wikidata and the English Wikipedia, consists of nearly five million entities and twenty million triplets covering a wide range of encyclopedic knowledge, mostly about people, places, and things such as movies.

Wikidata5M has two versions of dataset splits, i.e., a transductive split where models are evaluated for their performance in predicting missing links between known nodes (entities) that have been seen during training, and an inductive split where nodes for evaluation are unseen during training.
Conventional KGE approaches are generally designed for transductive evaluation, thus unsuitable for inductive evaluation.
As Wikidata5M is much larger than standard KG completion datasets (e.g., FB15K-237 and WN18RR in \autoref{tab:datasets}), 
a prohibitive increase in parameter size limits those traditional approaches to be applicable for the transductive setting here.

We report results of ranking metrics, i.e., mean reciprocal rank (MRR) and Hits@k, $k\in{1, 3, 10}$ under the standard filtered setting~\cite{bordes2013transe}. When there are ties during ranking, we treat the ground truth as the last in the tie, which is the worst-case ranking result as done in \citet{sun-etal-2020-evaluation}.\footnote{Ties are rare (fewer than 5\%) in our model predictions.}

\begin{table*}[tb]
\centering
\setlength{\tabcolsep}{3pt}
\begin{adjustbox}{max width=\textwidth}
\begin{tabular}{lrrrrr}
\toprule
\multirow{2}{*}{Model}  & \multirow{2}{*}{Params} & \multirow{2}{*}{MRR$\uparrow$} & \multicolumn{3}{c}{Hits$\uparrow$} \\\cmidrule(r{4pt}){4-6}
      & & & @1 & @3 & @10 \\\midrule
TransE~\cite{bordes2013transe}   & 2.4T & .253 & .170 & .311 & .392 \\
DistMult~\cite{yang2014distmult}  & 2.4T & .253 & .208 & .278 & .334 \\
ComplEx~\cite{trouillon2016complexe}  &2.4T & .281 & .228 & .310 & .373 \\
SimpIE~\cite{kazemi2018simple}  &2.4T & .296 & .252 & .317 & .377 \\
RotatE~\cite{sun2018rotate}   &2.4T & .290 & .234 & .322 & .390 \\ \midrule
KEPLER~\cite{wang2021KEPLER}   & 110M & .210 & .173 & .224 & .277\\ 
MLMLM~\cite{clouatre-etal-2021-mlmlm}   & 355M & .223 & .201 & .232 & .264\\ 
KGT5~\cite{saxena-etal-2022-sequence}   & 60M & .300 & .267 & .318 & .365\\ 
KGT5-ComplEx Ensemble~\cite{saxena-etal-2022-sequence}   & 674M & .336 & .286 & .362 & .426\\ 

\hitctx (Ours)    & 110M & \textbf{.377} & \textbf{.332} & \textbf{.398} & \textbf{.456} \\
\bottomrule
\end{tabular}
\end{adjustbox}
\caption{Comparison between the proposed method and baseline methods on Wikidata5M under the transductive setting. Results of transductive KGE methods (first section) are taken from \citet{wang2021KEPLER}. Numbers in \textbf{bold} represent the best results.}
\label{tab:transductive_results}
\end{table*}
\begin{table}[tb]
\centering
\setlength{\tabcolsep}{3pt}
\begin{adjustbox}{max width=\columnwidth}
\begin{tabular}{lrrrr}
\toprule
\multirow{2}{*}{Model}   & \multirow{2}{*}{MRR$\uparrow$} & \multicolumn{3}{c}{Hits$\uparrow$} \\\cmidrule(r{4pt}){3-5}
      & & @1 & @3 & @10 \\\midrule
MLMLM~\cite{clouatre-etal-2021-mlmlm} & .284 & .226 & .285 & .348 \\
KEPLER~\cite{wang2021KEPLER}   & .402 & .222 & .514 & .730\\ 
BLP-SimplE~\cite{daza2021inductive}   & .493 & .289 & .639 & .866 \\

\hitent (Ours)    & \textbf{.634} & \textbf{.517} & \textbf{.703} & \textbf{.871} \\
\bottomrule
\end{tabular}
\end{adjustbox}
\caption{Link prediction results on the inductive split of Wikidata5M. Numbers in \textbf{bold} represent the best results.}
\label{tab:inductive_results}
\end{table}
\subsection{Experimental Settings}

We adopt a nine-layer entity Transformer and a three-layer context Transformer.
Each layer is parameterized in the same way of BERT\textsubscript{BASE}~\cite{devlin-etal-2019-bert}.
We use the entity descriptions released with Wikidata5M as the entity surface forms, tokenized by BERT's uncased subword tokenizer, with a maximum length of 27.
We randomly sample five neighbor nodes of the source entity from the training graph as the graph context for the transductive setting.
In the inductive setting, we cannot make use of this graph context because the nodes used in evaluation are disconnected to the training graph.
Although we observe that including such context can help for the transductive case,
we leave how to incorporate graph neighborhood for the inductive setting to future work.\footnote{\citet{markowitz-etal-2022-statik} sample a set of nodes from the test graph as the graph context during evaluation. We do not adopt this practice since it is not standard and cannot compare to results in other works.}
To distinguish the two \hit models trained with and without graph contexts, we call the former \hitctx and the latter single-entity model \hitent.

We use the AdamW~\cite{loshchilov2018decoupled} optimizer with a learning rate of 5e-4.
For models initialized with pre-trained BERT weights, we set a lower learning rate of 3e-5 for the loaded weights.
Learning rates linearly warm up for the first 10\% of total steps, then decay to zero.
We allocate a batch of 128 examples to each GPU with an effective total batch size of 2048 on 16 GPUs.
The in-batch negative sample size for each GPU is set to 1000.
We run each model once with a fixed set of seeds.
Training for 10 epochs usually takes about 60 hours on 16 NVIDIA Tesla V100 GPUs using PyTorch's DDP trainer. Evaluation is done with the last epoch checkpoints in an RTX 2080.
Details of hyperparameter tuning are described in \autoref{sec:expdetails}.

\subsection{Experimental Results}
We compare our \hit model against two groups of baselines:
(1) conventional KGE methods such as TransE~\cite{bordes2013transe}, DistMult~\cite{yang2014distmult}. These methods store entity embeddings in a large lookup table and use simple relation-aware scoring functions to compute the similarity between entities.
As mentioned earlier, the size of the lookup table accounts for the large parameter size of these methods, and they only work for the transductive setting;
(2) compositional knowledge representation approaches, including KEPLER~\cite{wang2021KEPLER}, MLMLM~\cite{clouatre-etal-2021-mlmlm}, and KGT5~\cite{saxena-etal-2022-sequence}.
These methods all use a Transformer to encode entity surface forms, but they differ in the way they decode target entities.
MLMLM and KGT5 both leverage distribution of the language modeling objective to estimate the likelihood of target entities, while KEPLER uses a TransE-like scoring function.
In contrast, we use the context Transformer as a decoder to directly predict the target entity.

Our \hit model outperforms all prior methods by a large margin, under both the transductive (Table~\ref{tab:transductive_results}) and inductive (Table~\ref{tab:inductive_results}) settings on Wikidata5M, establishing a new state-of-the-art result.
In the transductive setting, we see a clear advantage of compositional knowledge representation approaches over traditional KGE methods in terms of model size.
However, previous compositional knowledge representation approaches are still on par with best performing KGE methods such as SimpIE in terms of link prediction performance, demonstrating the difficulty in building compositional representation from entity surface forms.
\hit surpasses all baselines and even beats the ensemble of a T5-like Transformer model and transductive ComplEx embeddings.

\subsection{Ablations and Analyses}
\label{sec:ablations}
To better understand the design choices we made and what contributes to the performance, we further provide ablations and analyses.
Due to computational resource constraints, all ablation and analytical experiments are run for five epochs with other training details the same as in the main experiments.
As shown in Table~\ref{tab:ablations}, the full models retrained for five epochs are only slightly worse.

\paragraph{Support Set}
Support sets consist of a source entity's neighbors in the training graph. This compositional feature relieves the model from the burden of memorizing the local graph context of the source entity~\cite{chen-etal-2021-hitter}, and could help the model identify entities when their surface forms are ambiguous.
The superior performance of \hitctx over \hitent on the transductive setting demonstrates the helpfulness of support sets.
More evidence will be provided in Section~\ref{sec:transfer} on the transfer experiments.
Since the entities in the inductive setting are disconnected from the training graph, we cannot use support sets in the inductive setting.

\begin{table}[t]
\centering
\begin{adjustbox}{max width=\linewidth}
\begin{tabular}{lrrrr}
\toprule
\multirow{2}{*}{\normalsize{Model}} & \multicolumn{2}{c}{Inductive} & \multicolumn{2}{c}{Transductive} \\\cmidrule(r{4pt}){2-3} \cmidrule(l){4-5}
                       & MRR           & H@10          & MRR          & H@10        \\\midrule
\hitctx & - & - & 36.4 & 44.1 \\
\hitent & 61.2 & 84.4 & 34.6 & 42.0 \\
Early fusion & 62.9 & 85.0 & 36.9 & 44.8 \\
Random init & 52.5 & 78.0 & 33.3 & 40.7 \\
Entity name & 25.0 & 40.1 & 29.0 & 36.6 \\
\bottomrule
\end{tabular}
\end{adjustbox}
\caption{Results of different model variants on dev sets of Wikidata5M. Models trained under the inductive setting do not have a support set of neighbor entities.}
\label{tab:ablations}
\end{table}

\paragraph{Entity Transformer Initialization}
To better utilize the compositional information of entity surface forms, we initialize the entity Transformer with pretrained BERT weights.
As anticipated, the model performance improves when using pretrained BERT weights compared to random initialization.

\paragraph{Early/Late Fusion of Relation Embeddings}
Recall that in \hit design, all the relation embeddings (either from predicate or neighbors) are directly fed into the context Transformer, skipping the entity Transformer.
It is possible that the entity Transformer could benefit from the relational information.
The ablation results show that this is indeed the case, but the improvement is not significant.
We view this as a trade-off between the expressive power of the entity Transformer and computational efficiency as the entity Transformer has a much larger input size than the context Transformer and thus adding one more token to the input would increase the $O(N^2)$ computation cost more.

\paragraph{Entity Surface Forms}
When replacing conventional lookup entity embeddings with compositional entity embeddings produced from entity surface forms, the choice of the type of surface forms and the corresponding encoder is conceivably important for modeling meaningful representations of entities.
As we can see in Table~\ref{tab:ablations}, when using entity names, the model performance on link prediction decreases drastically.
Since entity names are generally shorter, more memorization and less generalization of the model are expected.
This also coincides with the observation that entity names are not as harmful in the transductive setting, where the model can see and memorize all entity names in the training graph.

\begin{table}[t]
\centering
\begin{adjustbox}{max width=\linewidth}
\begin{tabular}{lrr}
\toprule
Model & MRR           & H@10                 \\\midrule
T6B6 & 59.4 & 83.4 \\
T5B7 & 60.1 & 83.8 \\
T4B8 & 61.1 & 84.2 \\
\midrule
NS 100 & 52.4 & 78.2 \\
NS 400 & 58.7 & 83.1 \\
NS 700 & 60.3 & 84.0 \\
\midrule
\hitent & 61.2 & 84.4 \\
\bottomrule
\end{tabular}
\end{adjustbox}
\caption{Results of the Shallow Entity Transformer and Negative Sample Size experiments on the dev set of Wikidata5M's inductive split. TnBm stands for \hit with a n-layer context Transformer and a m-layer entity Transformer.}
\label{tab:ablations2}
\end{table}
\paragraph{Shallow Entity Transformer}
From Table~\ref{tab:ablations2}, we can see consistent improvements coming from increasing the proportion of bottom entity Transformer layers in the overall architecture, backing our design choice of the deep entity Transformer, and shallow context Transformer architecture.\footnote{Further increases of entity Transformer size are difficult due to GPU memory constraint as entity Transformer usually has a much larger input size than that of the context Transformer.} This further suggests that encoding entity surface forms requires non-trivial modeling expressiveness.

\begin{table*}[]
\centering
\begin{adjustbox}{max width=\textwidth}
\begin{tabular}{lcccccccc}
\toprule
 & \multicolumn{4}{c}{Transductive} & \multicolumn{4}{c}{Inductive} \\\cmidrule(r{3pt}){2-5} \cmidrule(){6-9}
 & \multicolumn{2}{c}{FB15K-237} & \multicolumn{2}{c}{WN18RR} & \multicolumn{2}{c}{FB15K-237} & \multicolumn{2}{c}{WN18RR} \\\cmidrule(r{3pt}){2-3} \cmidrule(){4-5} \cmidrule(l{3pt}r{3pt}){6-7} \cmidrule(){8-9}
Model  &      MRR         &     H@10         &    MRR  &    H@10  &      MRR         &     H@10         &    MRR  &    H@10       \\\midrule
 HittER~\cite{chen-etal-2021-hitter} & 37.3 & 55.8 & 50.3 & 58.4 & - & - & - & - \\
 BLP-TransE~\cite{daza2021inductive} & - & - & - & - & 19.5 & 36.3 & 28.5 & 58.0 \\\midrule
\hitent & 34.5 & 53.0 & 44.8 & 67.5 & 23.0 & 39.1 & 30.9 & 51.4 \\
\hitent-WD     & 37.6 & 56.2 & 52.1 & 74.0 & 27.5 & 44.7 & 39.5 & 60.5 \\\midrule
\hitctx   &  34.4 & 52.8 & 50.0 & 71.0 & - & - & - & - \\
\hitctx-WD     & 37.4  & 56.3 & 55.0 & 74.9 & - & - & - & - \\
                        \bottomrule
\end{tabular}
\end{adjustbox}
\caption{Our test results on two regular-scale datasets for transductive and inductive KG completion. The first part of the table presents the results of the state-of-the-art models from previous work, while the second/third parts show the results of our \hit models using different initialization methods. WD stands for Wikidata5M pretraining.}
\label{tab:transfer}
\end{table*}
\paragraph{Negative Sample Size}
Negative sampling is widely adopted in KG representation learning to distinguish the right target entity from others in the intractable entity space during training. It has been shown to have a huge impact on the quality of learned representations~\cite{kotnis2017analysis}.
The bottom part of Table~\ref{tab:ablations2} shows that increasing the number of negatives from 100 to 1000 consistently improves link prediction results and further improvements seem to be marginal.

\section{Transfer Experiments}
\label{sec:transfer}
In this section, we transfer Wikidata5M pre-trained models to two other KGs by continual training and evaluate the \emph{out-of-distribution} adaptability of \hit.

\subsection{Datasets}
We use the FB15K-237 dataset~\cite{toutanova-chen-2015-observed} and the WN18RR dataset~\cite{dettmers2018conve} as the target KGs.
As another encyclopedic KG, FB15K-237 comes from arguably the same domain as Wikidata5M, 94\% of the entities in FB15K-237 are also in Wikidata5M, but they focus on different relation sets.
We observe that about 80\% of the unique $(e_s, e_o)$ pairs in FB15K-237 are not present in Wikidata5M, suggesting a distributional shift.
WN18RR is a subset of WordNet~\cite{miller1995wordnet} that contains facts about word relations, such as synonyms, antonyms, hypernyms, etc., which presents a much larger domain shift than FB15K-237.
Besides the difference in domains, we also want to evaluate the ability of \hit to deal with entities unseen in the training graph.
To this end, we additionally adopt inductive versions of the two datasets~\cite{daza2021inductive}, where all triplets in the dev and test sets contain at least one novel entity.
Details of the datasets are provided in Table~\ref{tab:datasets}.

\subsection{Experimental Settings}
The experimental settings are mostly the same as in the previous section with the following exceptions.
Here, besides using an initialization with pre-trained BERT weights, we have another option to initialize the model with \hit model pre-trained on Wikidata5M.
When using this option, all weights are initialized with the pre-trained \hit model, except for the relation embeddings, which are randomly initialized.
We denote this model \hit-Wiki to distinguish it from the model initialized with BERT weights.
The entity surface forms for FB15K-237 and WN18RR are provided by \citet{daza2021inductive}, used by both inductive and transductive settings.
As the entity surface forms for WN18RR are generally shorter than those for encyclopedic knowledge, we cap the maximum number of subwords to 12 for WN18RR, which consequently allows us to use a larger negative sampling size of 1500.
We train the model for five epochs.

\subsection{Experimental Results}
As we can see from the results in Table~\ref{tab:transfer}, models initialized with Wikidata5M pre-trained checkpoints consistently outperform models initialized with pre-trained BERT weights across the board.

For both datasets, we observe more pronounced improvements for models initialized from pre-trained \hit under the inductive setting, which is likely more challenging than the transductive one.
In section~\ref{sec:ablations}, we have seen the benefit of support sets under the transductive setting of Wikidata5M.
Here we can see that the support sets are also helpful for the transductive setting of WN18RR but not FB15K-237, which is consistent with the observation in \citet{chen-etal-2021-hitter}.
We also note that randomly initialized models are hard to converge for these two datasets, suggesting that a good initialization is critical for smaller datasets given the instability of large Transformer models.

\begin{figure}[t]
\centering
\includegraphics[width=0.9\linewidth]{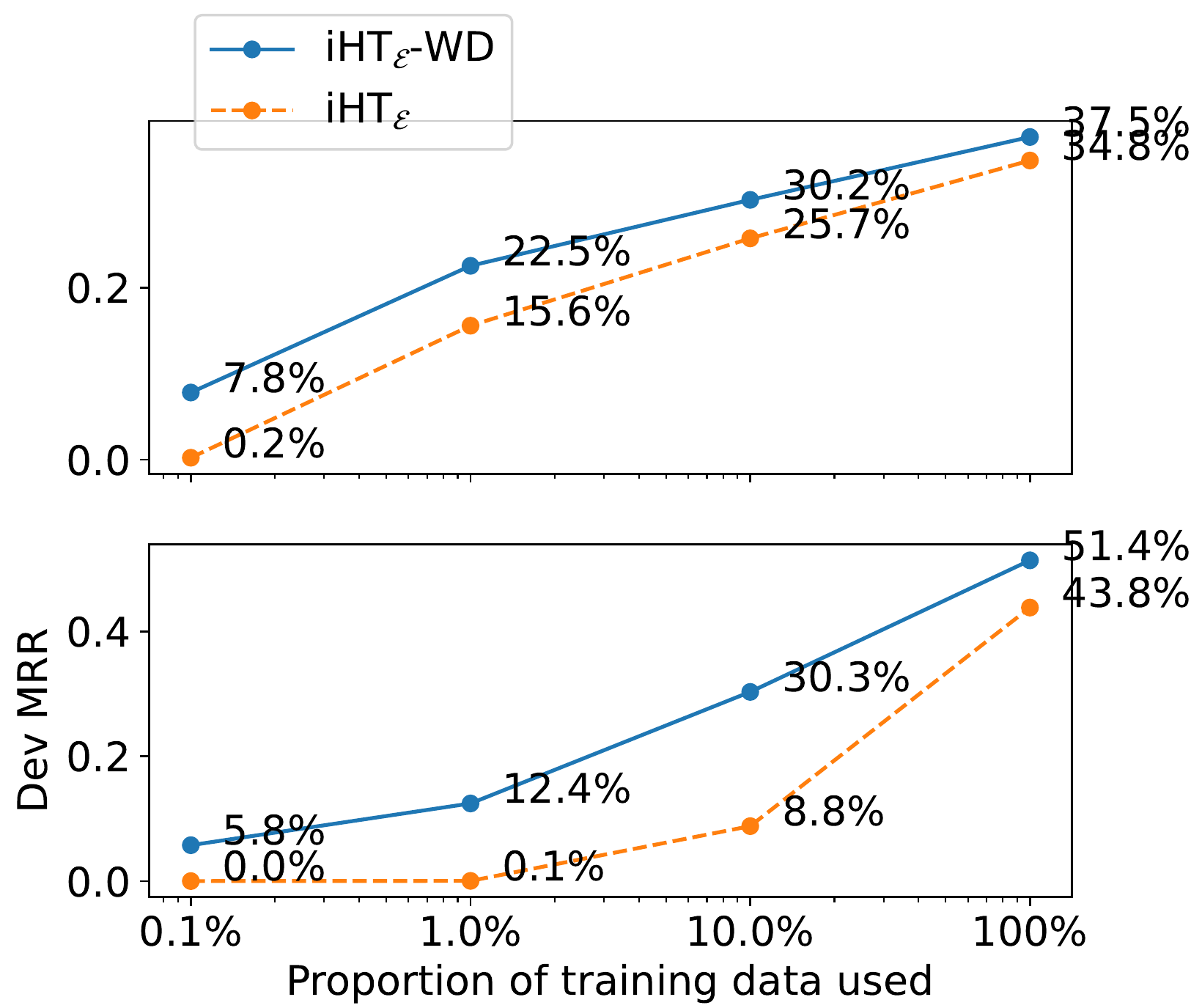}
\caption{Transfer performance of \hitent-WD (large-scale pretrained) and \hitent (BERT initialized) trained on FB15K-237 (top) and WN18RR (bottom) with different training data sizes.}
\label{fig:transfer_data_scale}
\end{figure}

To investigate the transferability of \hit in low-resource scenarios, we further conduct experiments on the standard transductive setting of FB15K-237 and WN18RR with different training data sizes.
From Figure~\ref{fig:transfer_data_scale}, we can see that the performance of the model scales almost logarithmically with the training data size.
The pre-training on Wikidata5M (inductive split) provides consistent improvements over the model initialized with only BERT weights.
With only 10\% of the training data, the Wikidata5M pre-trained model is able to achieve 70-86\% of the performance of the BERT-initialized model trained on the full training set, suggesting that the large-scale pre-training helps to reduce the need for more training data.

\section{Related Work}

Conventional approaches to KG completion focus on learning a low-dimensional embedding for each entity and relation based purely on the structure of KGs~\cite{wang2017knowledge,ji2021survey}.
The generalizability of such transductive approaches is limited within learning new relations between a pre-defined set of entities that appear during training.
Moreover, as they heavily rely on the embeddings of entities to store relational knowledge, they are not scalable to large-scale KGs without excessive memory consumption and engineering efforts~\cite{lerer2019pytorch}, which further limits the amount of knowledge they can access.

To address the above limitations, inductive approaches abandon the notion of entity embeddings and instead represent entities with their textual information~\cite{wang-etal-2014-knowledge,xie2016representation} or rely purely on the multi-relational structure of KGs to perform rule induction~\cite{teru2019inductive}, thus enabling the generalization to new entities that are not connected to the training graph.
Nevertheless, the extended generalizability is still limited to a single KG.

Inspired by the success of Transformer in transfer learning~\cite{devlin-etal-2019-bert}, Transformer-based KG completion models are proposed to capture the compositional semantics of entities and utilize the transferable knowledge in pre-trained LMs~\cite{yao2019kgbert,bosselut-etal-2019-comet,daza2021inductive,clouatre-etal-2021-mlmlm}.
A long line of work considers jointly training a Transformer model with both KG completion and language modeling objectives~\cite{wang2021KEPLER,yasunaga2022deep}.
\hit keeps the notion of entity representation for relational scoring as opposed to modified token-based scoring~\cite{clouatre-etal-2021-mlmlm,saxena-etal-2022-sequence} to facilitate potential application with entity representations.
\hit can be initialized with pre-trained LMs like BERT but further training is only performed on the KG completion task.

Transferring knowledge between different KGs is less explored in literature, largely focused on KGs with shared components that can be aligned~\cite{sun13benchmarking}, such as multilingual KGs~\cite{chen-etal-2020-multilingual}.
Most closely related to the concept of large-scale pre-training on KGs, \citet{kocijan-lukasiewicz-2021-knowledge} adopts an RNN to encode textual representations of entities and conventional relational scoring functions such as \citet[TuckER]{balazevic-etal-2019-tucker} for KG completion.
Their encoder is first pre-trained on a large knowledge base, and then fine-tuned to generate entity representations on target knowledge bases.
The pre-training improves model performance on small-scale OKBC datasets but not on standard KG completion datasets such as FB15K-237 and WN18RR.
\hit employs a much more expressive Transformer model and offers consistent improvements.

\section{Conclusion}
We present \hit, a Transformer model for compositional knowledge graph representation learning.
When evaluated on the KG completion task, \hit achieves state-of-the-art results on Wikidata5M for both inductive and transductive settings.
Even given such a large-scale knowledge graph containing millions of entities and triplets, \hit can be successfully trained with regular memory footprints.
We further show that \hit learns transferable representations from Wikidata5M, which is able to boost the model performance on other knowledge graphs from different domains.

\section{Limitations}
\label{sec:limitations}
As an inductive approach, \hit relies on the textual information of entities to learn their representations.
Despite the wide availability of textual representations for popular KGs such as encyclopedic ones like Wikidata, \hit is not applicable to KGs that do not have textual information for entities and it can suffer from poor quality of textual information as demonstrated in the ablation study replacing entity descriptions with shorter entity names (Table~\ref{tab:ablations}).
\hit keeps the notion of entity representation to support potential entity-centric applications, but it is consequently slower during inference than the token-based scoring methods~\cite{clouatre-etal-2021-mlmlm,saxena-etal-2022-sequence} that do not encode entities separately at all.
Future work is needed to explore \hit's effectiveness in a more diverse set of KGs and downstream tasks beyond KG completion.

\bibliography{anthology,my}
\bibliographystyle{acl_natbib}

\appendix

\section{Dataset Details}
\label{sec:datasetdetails}
The FB15K-237 dataset~\cite{toutanova-chen-2015-observed} is under a Microsoft Research Data License Agreement which allows uses for non-commercial or research purposes.
The WN18RR dataset~\cite{dettmers2018conve} is under an MIT License.
The inductive versions\footnote{\href{https://github.com/dfdazac/blp}{https://github.com/dfdazac/blp}} of the two datasets~\cite{daza2021inductive} are also under an MIT License.
Wikidata5M\footnote{\href{https://deepgraphlearning.github.io/project/wikidata5m}{https://deepgraphlearning.github.io/project/wikidata5m}}~\cite{wang2021KEPLER} is publicly available but not clearly licensed.
Our research use of these datasets are aligned with their intended purposes.
Datasets derived from Freebase and Wikidata contain information directly associated with public figures.
Such information is largely non-private and moderated by the Wikimedia community.

\section{Experimental Setting Details}
\label{sec:expdetails}
We implement our proposed \hit model with PyTorch~\cite{paszke2019pytorch}.
We follow \citet{devlin-etal-2019-bert} to configure each of our Transformer layers and initialize the bottom entity transformer with pretrained BERT weights (\texttt{bert-base-uncased}) from huggingface's \texttt{transformers} library~\cite{wolf-etal-2020-transformers}.
KG completion evaluation is performed with the LibKGE~\cite{Ruffinelli2020You}.

We perform simple hyperparameter tuning on 
\begin{itemize}
    \item the number of negative samples used: \{ 100, 400, 700, 1000 \}
    \item layer combinations of the two Transformer blocks: \{ T6B6, T5B7, T4B8, T3B9 \}
\end{itemize}
These results are present in \autoref{tab:ablations2}. Other hyperparameters such as learning rates, learning rate scheduler, batch size, random seeds are manually determined without tuning.

\end{document}